\documentclass{article}
\usepackage{spconf,amsmath,graphicx}
\usepackage{todonotes}
\usepackage{booktabs}
\usepackage{subfigure}
\usepackage{romannum}
\usepackage{graphicx}
\usepackage{soul}
\usepackage{algorithm}
\usepackage[noend]{algpseudocode}
\usepackage{algorithmicx,algorithm}
\usepackage{multirow}
\usepackage[font=small,labelfont=bf]{caption}

\title{Speeding up convolutional networks pruning with coarse ranking} 

\name{Zi Wang$^{1,*}$, Chengcheng Li$^{1,*}$, Dali Wang$^{1,3}$, Xiangyang Wang$^2$, Hairong Qi$^1$
\thanks{$^*$ With equal contribution. Submitted to the 2019 IEEE International
Conference on Image Processing. Personal use of this material is permitted.
However, permission to reprint/republish this material for advertising or promotional
purposes or for creating new collective works for resale or redistribution
to servers or lists, or to reuse any copyrighted component of this work
in other works must be obtained from the IEEE.}
}
\address{$^1$Department of Electrical Engineering and Computer Science, University of Tennessee, USA \\
$^2$School of Mathematics, Sun Yat-Sen University, China\\
$^3$Oak Ridge National Laboratory, USA}
\begin{document}
\maketitle

\begin{abstract}
Channel-based pruning has achieved significant successes in accelerating deep convolutional neural network, whose pipeline is an iterative three-step procedure: ranking, pruning and fine-tuning. However, this iterative procedure is computationally expensive. In this study, we present a novel computationally efficient channel pruning approach based on the coarse ranking that utilizes the intermediate results during fine-tuning to rank the importance of filters, built upon state-of-the-art works with data-driven ranking criteria. The goal of this work is not to propose a single improved approach built upon a specific channel pruning method, but to introduce a new general framework that works for a series of channel pruning methods.  Various benchmark image datasets (CIFAR-10, ImageNet, Birds-200, and Flowers-102) and network architectures (AlexNet and VGG-16) are utilized to evaluate the proposed approach for object classification purpose. Experimental results show that the proposed method can achieve almost identical performance with the corresponding state-of-the-art works (baseline) while our ranking time is negligibly short. In specific, with the proposed method, $75\%$ and $54\%$ of the total computation time for the whole pruning procedure can be reduced for AlexNet on CIFAR-10, and for VGG-16 on ImageNet, respectively. Our approach would significantly facilitate pruning practice, especially on resource-constrained platforms.

\end{abstract}
\begin{keywords}
neural network acceleration, channel pruning, coarse ranking, deep neural network
\end{keywords}
\section{Introduction}
\label{sec:intro}

\begin{figure}[htb]
\centering
\begin{minipage}{0.9\columnwidth}
\centering
\includegraphics[width=\textwidth]{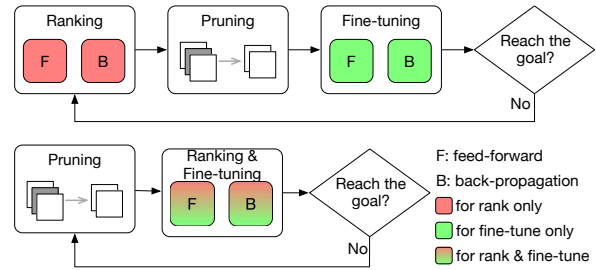}
\end{minipage}
\vskip -0.1in
\caption{Comparison of the typical channel pruning pipeline (top) and the proposed innovative pipeline (bottom).}
\label{fig:paradigm}
\vskip -0.05in
\end{figure}

The over-parameterization of deep convolutional neural networks (DCNN) has become a widely-recognized problem, while they have achieved significant success in a wide range of tasks \cite{iandola2014densenet,li2018fast,christiansen2018silico,Yuanchaodaen,wang2018deep,wang2019cellular}. In recent years, channel pruning \cite{han2015learning,li2016pruning,molchanov2016pruning,liu2017learning,he2017channel} has been proved to be an effective technique to reduce the size of a DCNN while sustaining its performance. 
It can remove the entire filters, as well as their corresponding feature maps \cite{li2016pruning,molchanov2016pruning,polyak2015channel}, so that no customized hardware are needed \cite{han2015learning,han2016eie}. A typical channel pruning procedure is as follows. Given a pre-trained DCNN, all the filters in the DCNN are ranked with a certain criterion and then the lowest ranked (i.e., least important) filters are pruned. Finally, the remaining sub-network is finetuned to alleviate performance loss. The ranking, pruning and fine-tuning are processed iteratively until a target (e.g., a desired pruning ratio or the maximal performance degradation) is satisfied, shown in Fig.~\ref{fig:paradigm}(top).
\vskip -0.1in
An appropriate criterion for ranking filters is considered as playing a dominant role in a successful channel pruning approach. Numerous criteria have been proposed \cite{li2016pruning,polyak2015channel,molchanov2016pruning,hu2016network}, among which, a significant branch is based on data-driven criteria. 
These approaches usually require a large number of training samples to be fed into the neural network in order to obtain reasonable ranking results, which can be computationally expensive. Our concurrent work shows that a precise filters ranking may not be necessary, since with a certain ranking criterion, pruning relative low-ranked filters can achieve almost identical performance as pruning the lowest-ranked ones. In this paper, we also verify this observation in the experiments (Section \ref{sec:coarse_sub}). Inspired by this finding, we propose a novel channel pruning approach based on coarse ranking, which embeds the ranking step into the fine-tuning step by utilizing the intermediate computation results during fine-tuning for ranking purpose. Consequently, ranking time can be significantly reduced. The proposed pipeline is illustrated in Fig.~\ref{fig:paradigm}(bottom).

We evaluate our approach with various benchmark architectures and datasets for object classification purpose. Experimental results demonstrate that the proposed approach can achieve comparable performance as the baseline, i.e. state-of-the-art approaches with precise ranking in a separate phase. Meanwhile, our ranking time can be reduced to less than $0.01$ second, which is negligibly short, and the overall pruning time is reduced by $75\%$ on AlexNet and around $60\%$ on VGG-16. These results demonstrate that our method can significantly speed up the channel pruning procedure, and can facilitate the pruning practice, especially on resource-limited platforms.


\section{Methodology}
\label{sec:method}
\subsection{The Overall Proposed Innovative Framework}
\label{sec:exp_method}
Given a pre-trained DCNN, most existing channel pruning approaches follow an iterative three-step pipeline shown in Fig.~\ref{fig:paradigm}(top): (1) ranking filters with an importance-based criterion, (2) pruning the least important filters, and (3) fine-tuning the remaining sub-network to alleviate performance loss. With this pipeline, most existing channel pruning methods are devoted to developing sophisticated ranking criteria. Among the numerous criteria, a significant branch is data-driven based criteria that calculate the importance of filters based on the values obtained in the feed-forward, back-propagation, or both passes of the neural network. In order to obtain a \emph{precise ranking} results, lots of training samples have to be fed into the DCNN and corresponding feed-forward and back-propagation are processed expensively. 

Even though we call step (1) as ranking, it actually includes two sub-steps, calculating the importance scores of filters with a certain criterion, and ranking the filters with their scores. What is computationally costly is not ranking a set of values but the process of feed-forward and back-propagation which are required for calculating the importance-based score of filters. Since the feed-forward and back-propagation are also processed during the fine-tuning phase, it is potential to borrow the computation results from fine-tuning instead of calculating them in a separate ranking phase. 

Hence, we propose to calculate the importance-based scores of filters with the intermediate computation results from the fine-tuning phase and use these scores to rank filters. We name this ranking approach as \emph{coarse ranking}. This approach would inevitably result in an imprecise ranking due to the dynamic network weights, which raises concerns that pruning performance would be degraded. However, based on our empirical study, a slightly imprecise ranking can actually achieve comparable performance with precise ranking. More experimental details are presented in Section \ref{sec:coarse_sub}. This finding provides a fundamental building block for the proposed innovative framework for channel pruning. The overall proposed framework is shown in Fig.~\ref{fig:paradigm}(bottom).
\subsection{Channel Pruning Methods with Data-driven Criteria}
\label{existingchannelpruning}
The proposed computationally efficient channel pruning framework works for a series of existing state-of-the-art methods using data-driven criteria for ranking. In the following paragraphs, we introduce several state-of-the-art channel pruning methods. We also present to use Spearman’s rank correlation coefficient \cite{zar1972significance} to measure the correlation between the baseline (precise ranking) and the proposed method (coarse ranking). 

{\bf Taylor expansion} \cite{molchanov2016pruning}. Consider a DCNN with $L$ convolutional layers, parameterized by $W=\{w_1^{1},w_1^{2},\cdots,w_1^{C_1},w_2^1,\\w_2^2,\cdots,w_2^{C_2},\cdots,w_L^{C_L}\}$, where $C_i$ is the number of filters in the $i$-th layer. Given a training set $D$, the loss function is $C(D,W)$. The objective of this pruning method is to minimize the loss change $|\Delta C|$ when there are at most $B$ non-zero filters in the network, parameterized by $W'$:
\vskip -0.2in
\begin{equation*}
    \underset{W'}{\text{minimize}}|C(D,W')-C(D,W)|~~s.t.~~||W'||_0 \leq B.
\end{equation*}
\vskip -0.08in
Suppose $H=\{h_1^1,h_1^2,\cdots,h_L^{C_L}\}$ are the feature maps of the corresponding filters $w$, then the loss change of removing $h_i$ can be calculated with Eq.~\ref{eq:loss_change}. 
\begin{equation}
    |\Delta C(h_i)| = |C(D,h_i=0) - C(D,h_i)|,
\label{eq:loss_change}
\end{equation}
where $C(D,h_i=0)$ and $C(D,h_i)$ are the losses when $h_i$ is and is not pruned, respectively. The first-order Taylor polynomial at $h_i=0$ is:
\vskip -0.2in
\begin{equation*}
    C(D,h_i=0) = C(D,h_i) + \frac{\delta C}{\delta h_i}h_i + R_1(h_i=0),
\end{equation*}
where the higher order residual $R_1(h_i=0)=\frac{\delta^2C}{\delta(h_i^2=\xi)}\frac{h_i^2}{2}$, and $\xi \in (0,h_i)$. Since this approach only consider the first order estimation, $R_1$ is neglected. Thus, Eq.~\ref{eq:loss_change} can be rewritten as:
\vskip -0.15in
\begin{equation*}
    |\Delta C(h_i)| = |C(D,h_i) + \frac{\delta C}{\delta h_i}h_i- C(D,h_i)| = |\frac{\delta C}{\delta h_i}h_i|.
\end{equation*}
$h_i$ and $\frac{\delta C}{\delta h_i}$ can be obtained in the feed-forward and back-propagation passes by feeding a batch of samples to the DCNN. $|\Delta C(h_i)|$ represents the importance of the filter corresponding to $h_i$.

{\bf Mean activation} \cite{polyak2015channel}. Suppose $a_i$ is the activation values of the feature map $h_i$ after ReLU. This approach considers the average $\ell_1$-norms of the associated feature maps, $A_i$, calculated by Eq.~\ref{eq:mean_activation} as the importance scores.
\vskip -0.1in
\begin{equation}
    A_i = \frac{1}{|a|}\sum_i a_i,
\label{eq:mean_activation}
\end{equation}
where $|a|$ is the dimensionality of the feature map.

{\bf Rankings correlation.} We use the Spearman's rank correlation coefficient \cite{zar1972significance} calculated with Eq. \ref{eq:corr} to measure the correlation between the proposed coarse ranking and the baseline (precise ranking). This coefficient takes values between $\left[ -1,+1\right]$, where $-1$ and $+1$ indicate absolute negative and positive correlations, respectively.
\begin{equation}
    r_s = 1 - \frac{6\times \sum_{i=1}^Nd_i^2}{N(N^2-1)},
\label{eq:corr}
\end{equation}
where $N$ is the number of filters and $d_i$ is the rank difference between the $i$-th filters of two ranking results.

\section{Experimental Results}

We evaluate the proposed approach with various benchmark network architectures and datasets, including AlexNet \cite{krizhevsky2012imagenet} on CIFAR-10 \cite{krizhevsky2009learning}, VGG-16 \cite{simonyan2014very} on Birds-200 \cite{WelinderEtal2010}, Flowers-102 \cite{nilsback2008automated}, and ImageNet \cite{deng2009imagenet} for image classification purpose. We verify the generality of the proposed approach with two widely-used channel pruning approaches with data-driven criteria for ranking, including Taylor expansion and mean activation, illustrated in Section \ref{existingchannelpruning}. Work \cite{molchanov2016pruning} demonstrated that Taylor expansion is the state-of-the-art work and outperforms mean activation significantly. Hence, without loss of generality, we mainly utilize Taylor expansion as the baseline and use mean activation as supplementary in the following experiments.
For AlexNet, we prune $10$ filters and finetune the sub-network for $100$ batches at each run; for VGG-16, we prune $50$ filters and finetune the sub-network for $500$ batches. For all cases, a learning rate of $0.0001$ with an SGD optimizer, and a batch size of 64 are used for fine-tuning. We run each experiment $5$ times and report the average results and associated standard deviation. Experiments are conducted with Pytorch \cite{paszke2017automatic} on $4$ NVIDIA Quadro P6000 GPUs and Intel Core i7-6850K CPU (3.60GHz).

\subsection{Study of Imprecise Ranking}
\label{sec:coarse_sub}
\begin{figure*}[htb]
\begin{center}
\centerline{
\subfigure[AlexNet, CIFAR-10]{
\includegraphics[width=0.475\columnwidth]{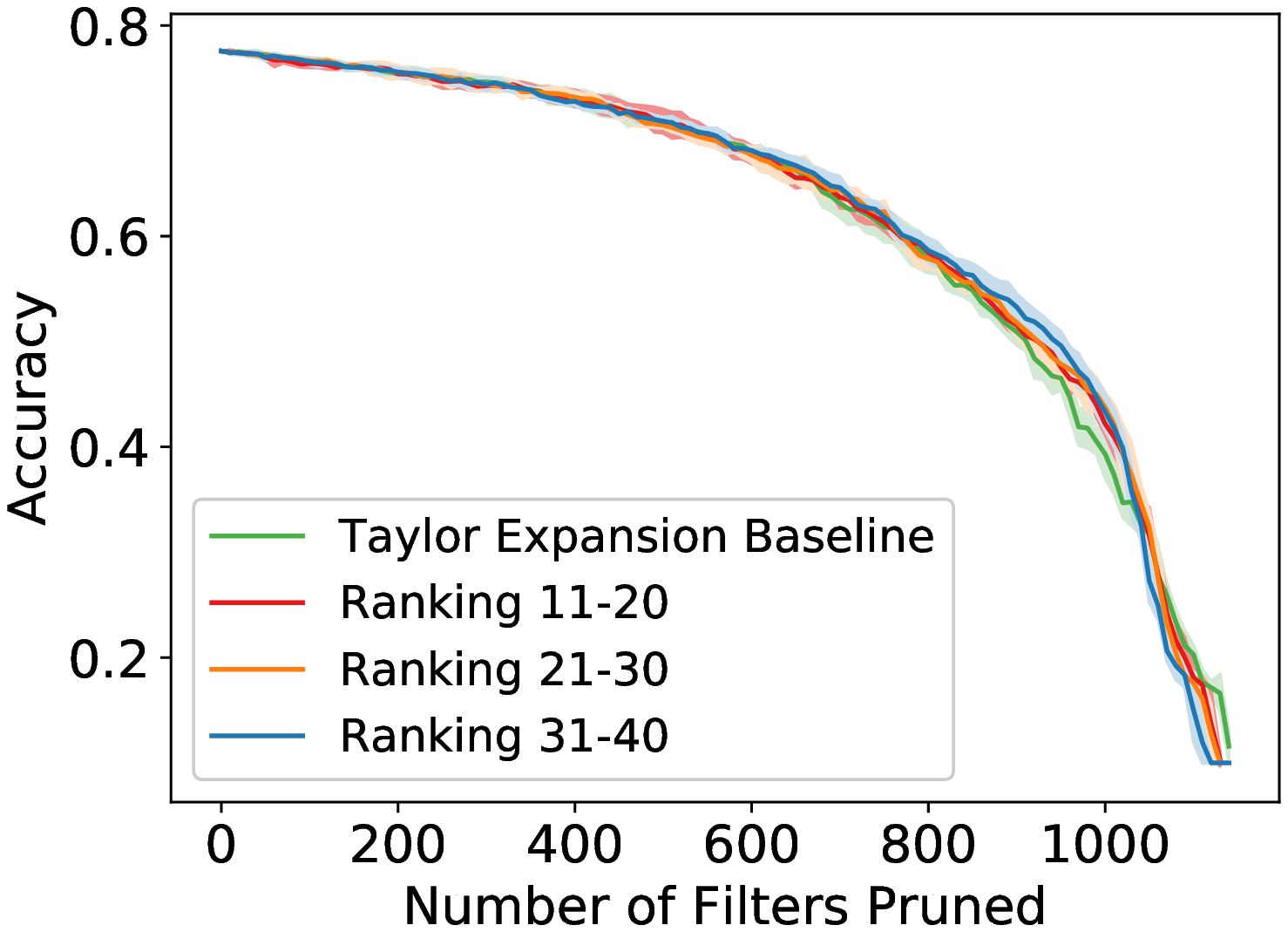}}
\subfigure[VGG-16, Birds]{
\includegraphics[width=0.475\columnwidth]{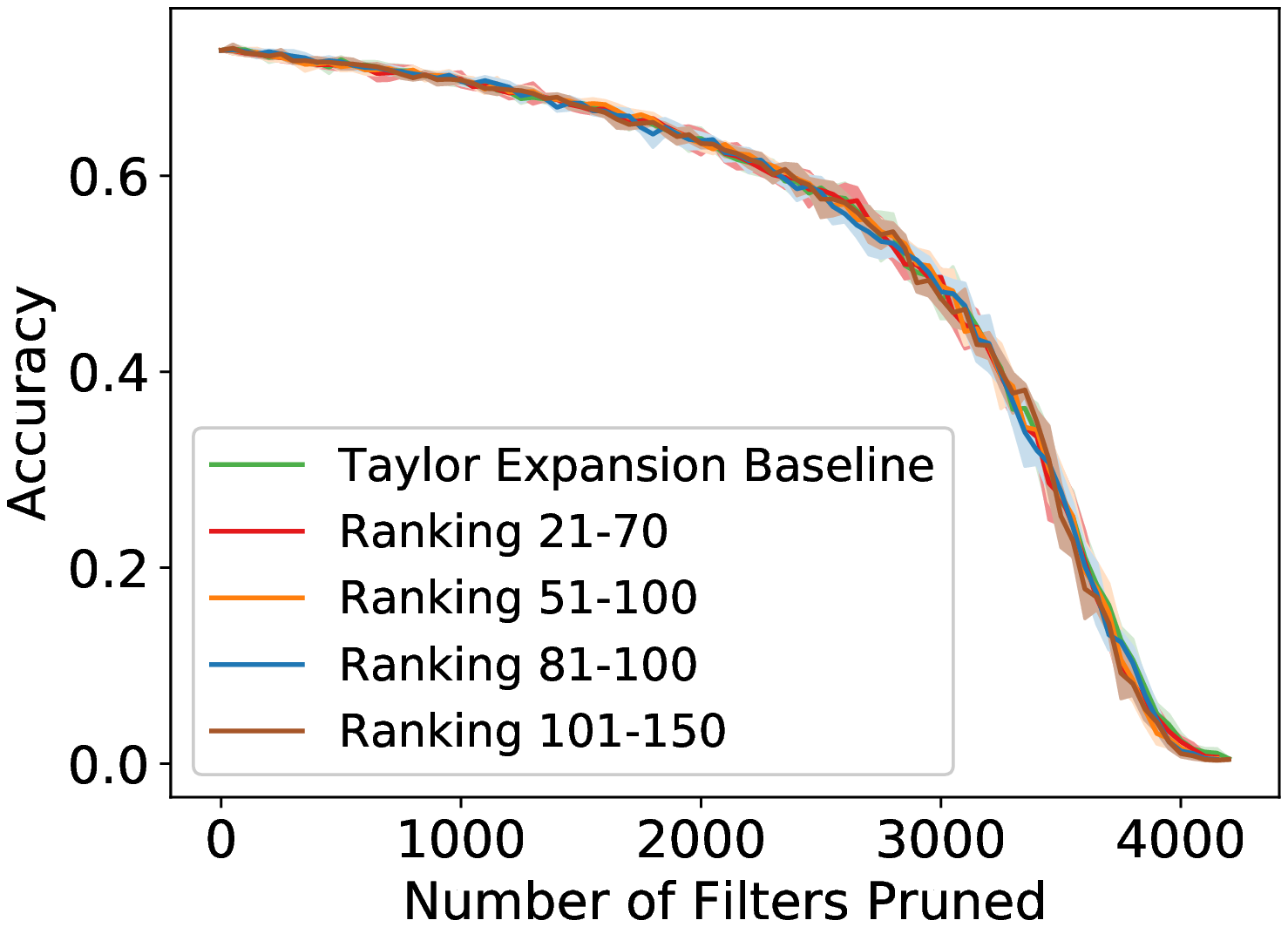}}
\subfigure[VGG-16, Flowers]{
\includegraphics[width=0.475\columnwidth]{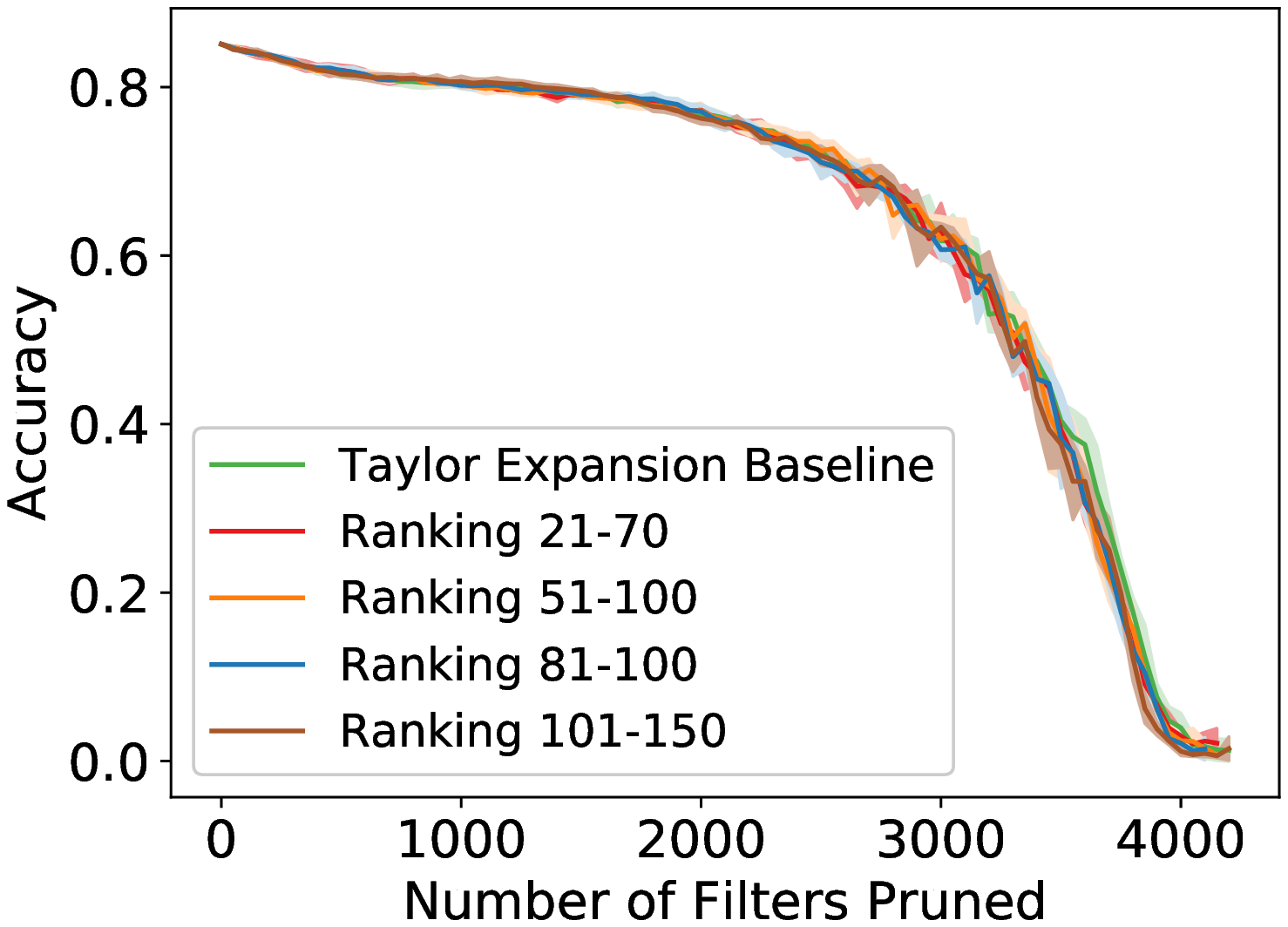}}
\subfigure[VGG-16, ImageNet]{
\includegraphics[width=0.475\columnwidth]{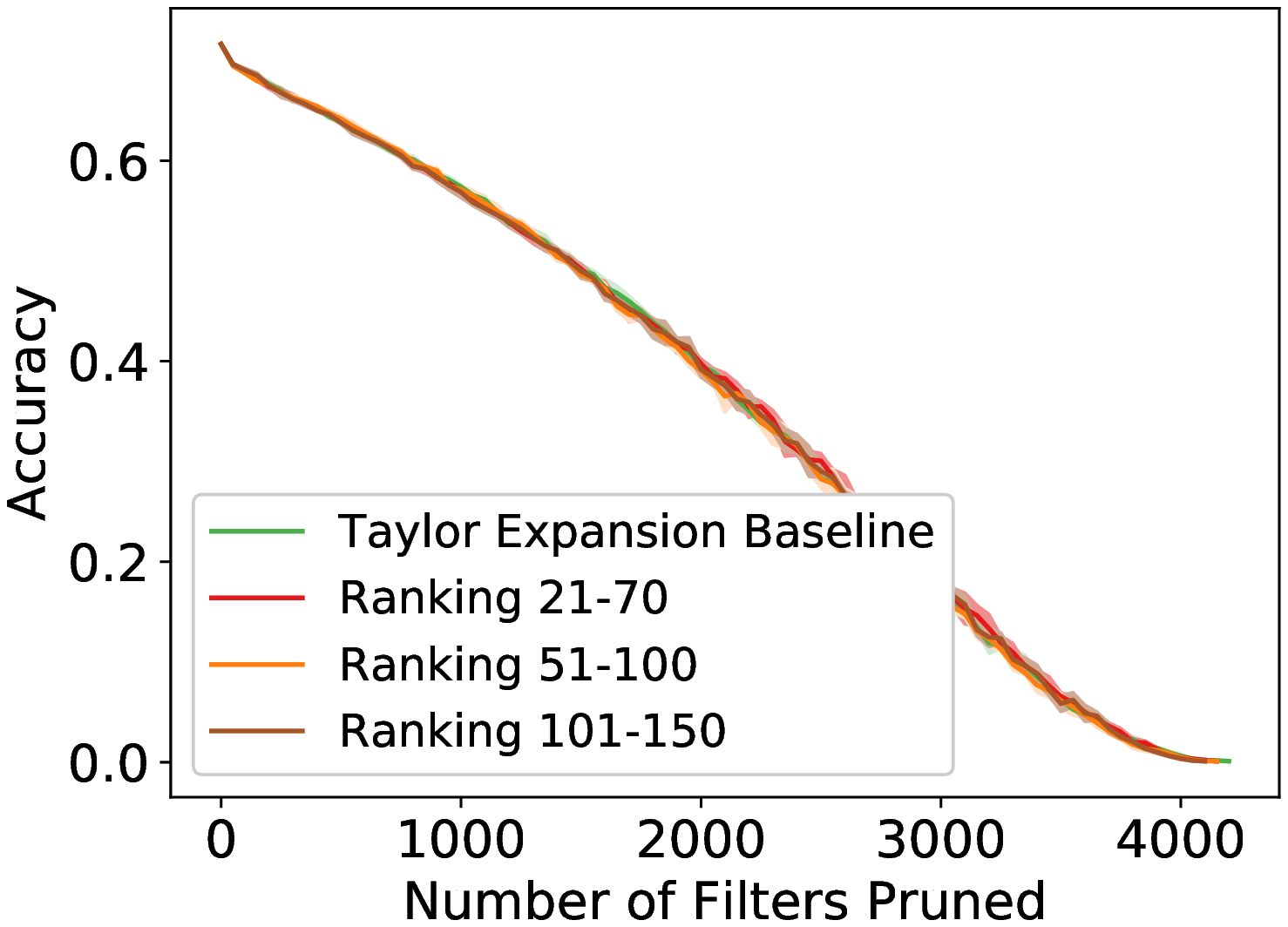}}
}
\vskip -0.17in
\caption{Performance comparison with different filter selection strategies on various architectures and datasets.}
\label{fig:relative_low}
\end{center}
\vskip -0.3in
\end{figure*}

Our proposed method is built upon the observation that pruning relatively less important filters results in comparable performance with pruning least important filters. Here we verify this finding with comparative study. In specific, we apply Taylor expansion-based criterion to rank filters. In each run of pruning, we compare the performance of different filters selection strategies, including pruning the filters with lowest ranks 1-10 (baseline), and those with relatively low ranks 11-20, 21-30, and 31-40 for AlexNet. Similarly, for VGG-16, the options are filters with ranks 1-50 (baseline), 21-70, 51-100, 81-130, and 101-150. All the other hyper-parameters and configurations remain identical for each pruning strategy. Experimental results are shown in Fig.~\ref{fig:relative_low}. It is observed that the performances of different filter selection strategies are almost identical with various structures and datasets. These results indicate that it is not necessary to strictly prune the lowest ranking filters, since pruning relatively low ranked filters results in comparable performance. Thus, the overall pruning procedure can be potentially accelerated with a coarse ranking strategy.

\subsection{Evaluation with Classification Accuracy}
In this subsection, we evaluate our proposed approach that embeds the filter ranking phase into the fine-tuning phase by borrowing the intermediate results of feed-forward and back-propagation from the fine-tuning phase. The Taylor expansion approach is still used as the baseline. In the first run of ranking, since the network has not been through a fine-tuning phase, we randomly select filters to be pruned. After that, we record the intermediate activation values and the corresponding gradients calculated in the feed-forward and back-propagation passes during fine-tuning and use the average results to calculate the importance scores of filters with the Taylor expansion criterion. Therefore, we avoid re-feeding the whole training set to the neural network and re-processing feed-forward and back-propagation in a separate ranking phase, which can save computation significantly. 

\begin{figure*}[ht]
\begin{center}
\vskip -0.1in
\centerline{
\subfigure[AlexNet, CIFAR-10]{
\includegraphics[width=0.475\columnwidth]{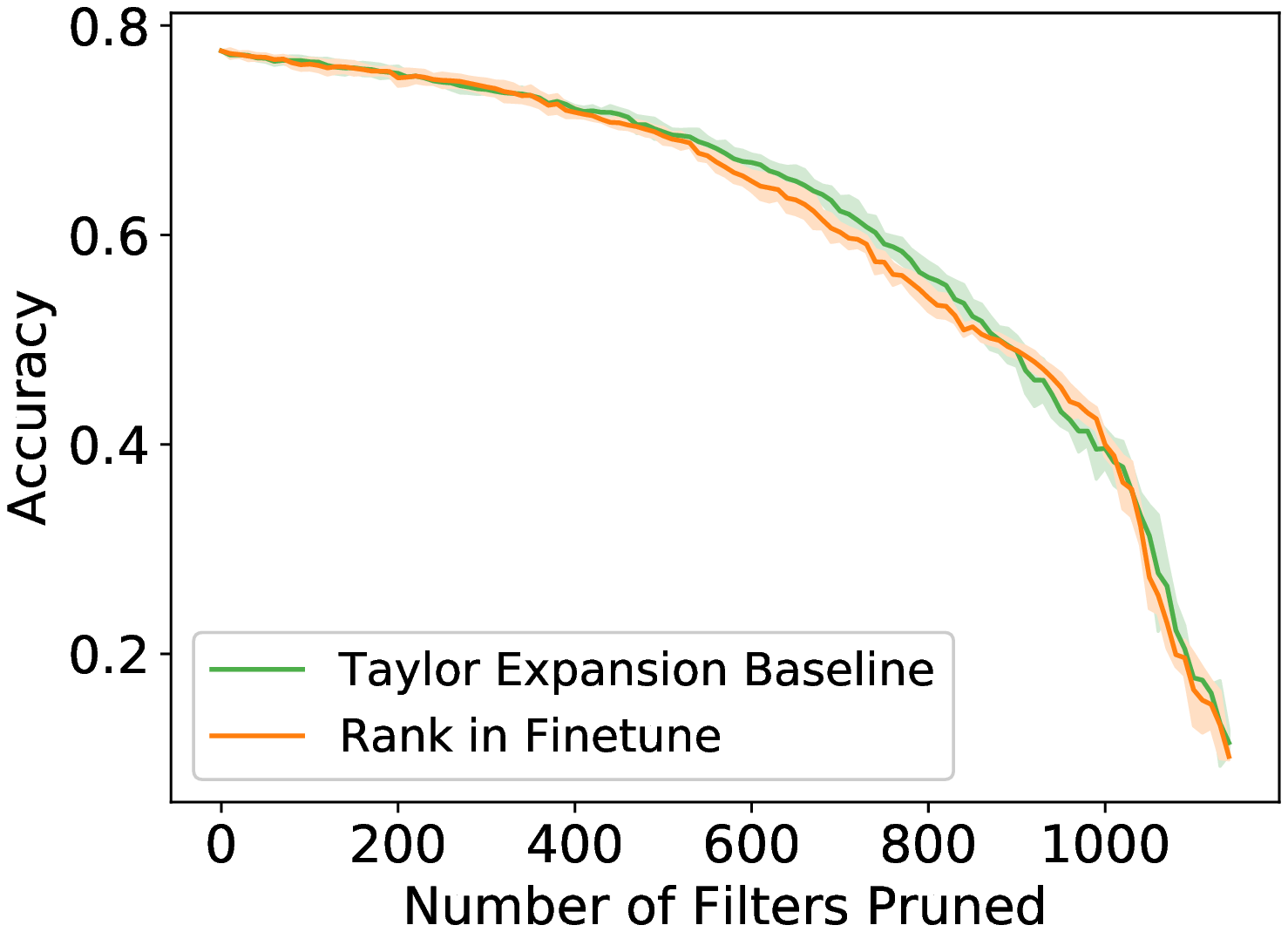}}
\subfigure[VGG-16, Birds]{
\includegraphics[width=0.475\columnwidth]{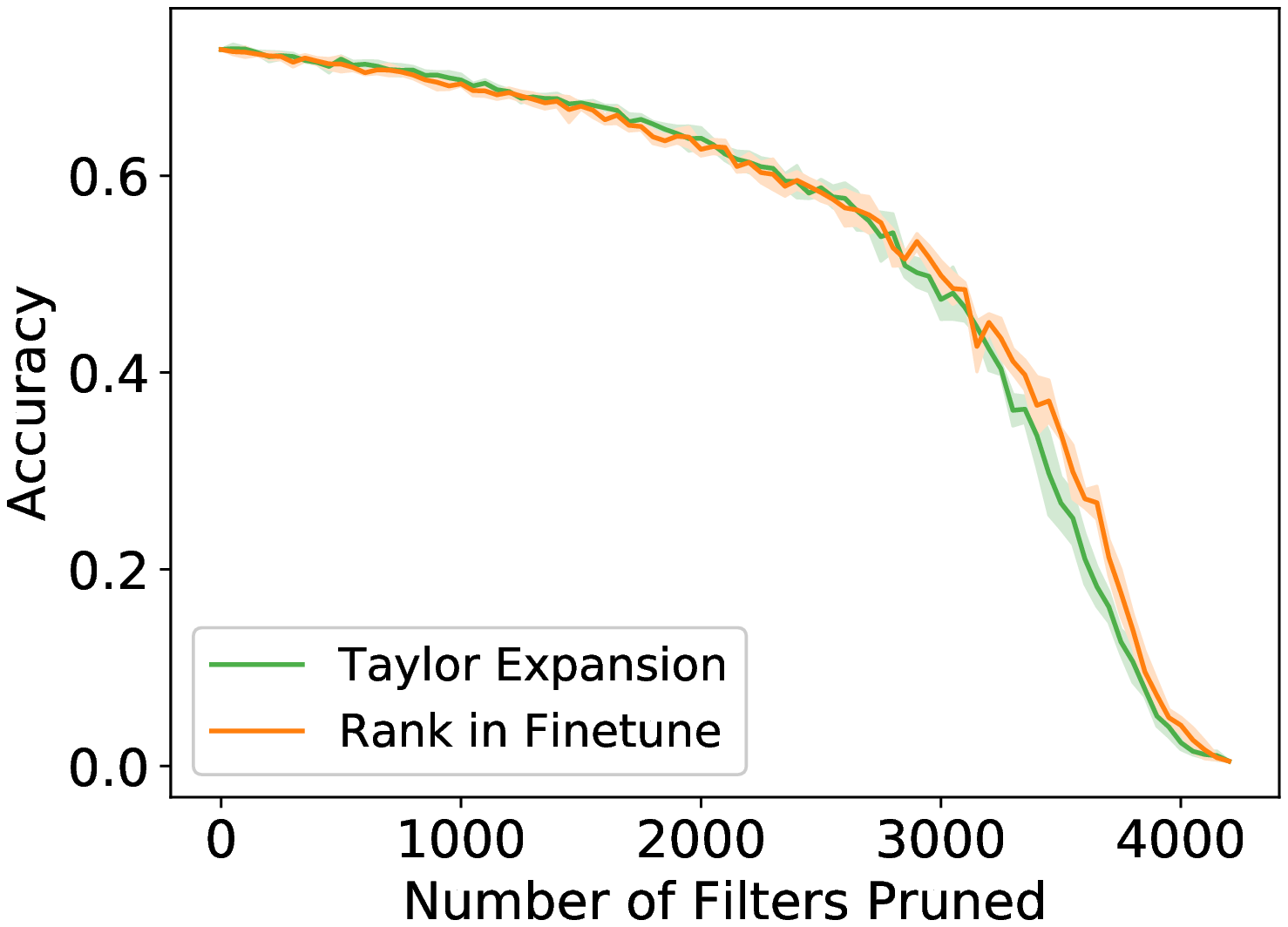}}
\subfigure[VGG-16, Flowers]{
\includegraphics[width=0.475\columnwidth]{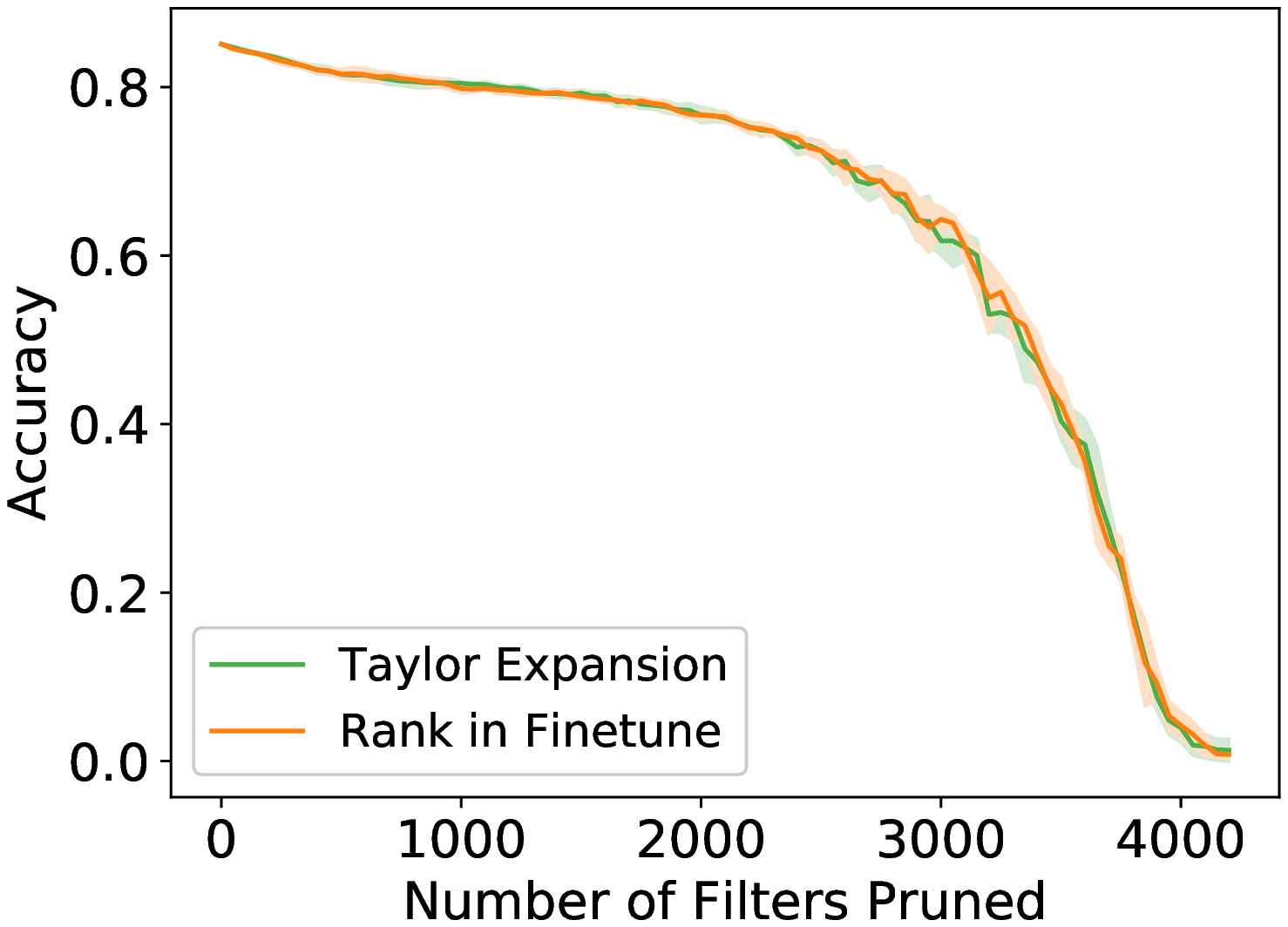}}
\subfigure[VGG-16, ImageNet]{
\includegraphics[width=0.475\columnwidth]{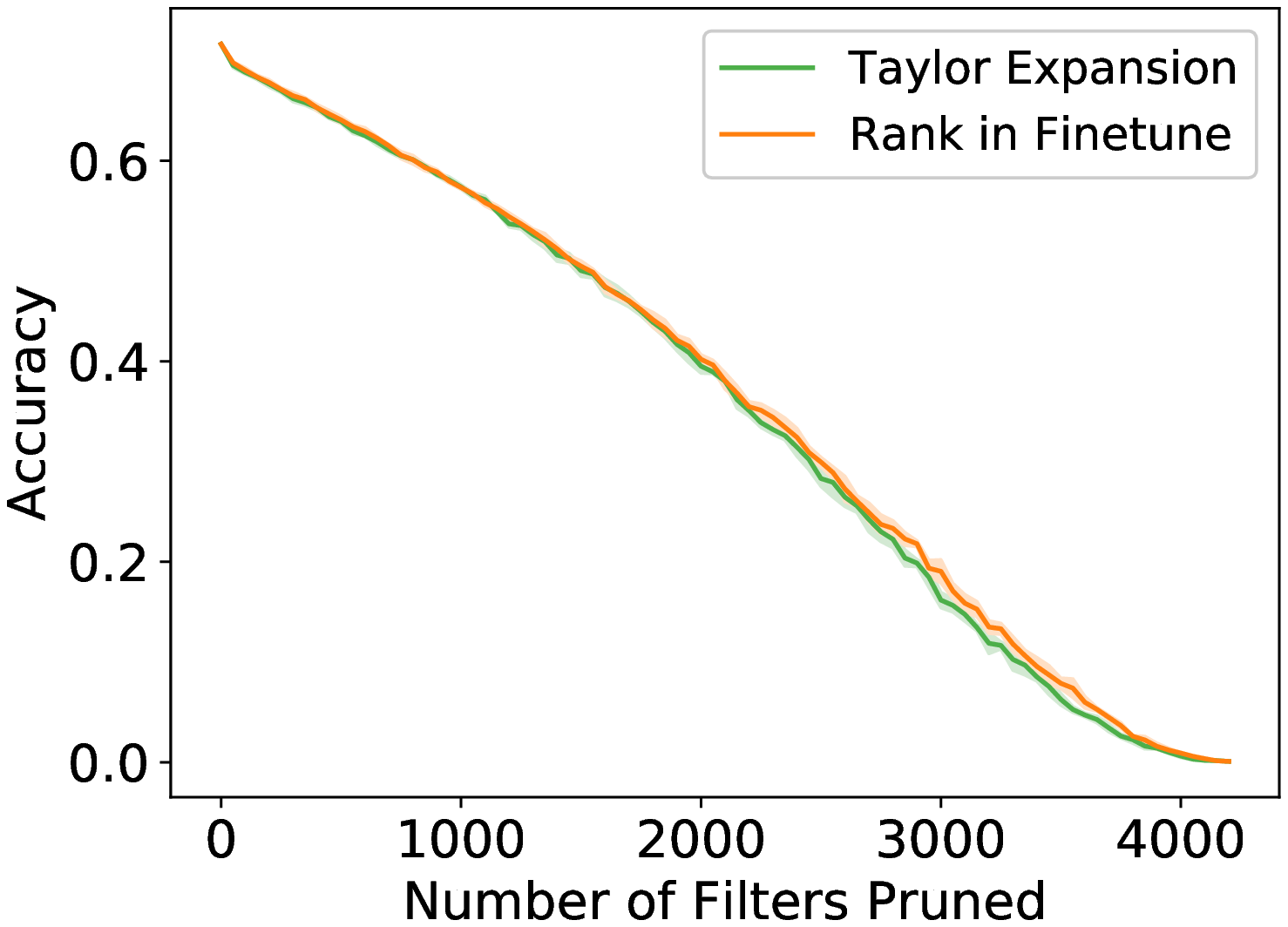}}
}
\vskip -0.17in
\caption{Performance comparison with the proposed approach (Rank in Finetune) and its corresponding baseline (Taylor expansion).}
\label{fig:coarse}
\end{center}
\vskip -0.37in
\end{figure*}

Experiment results in Fig.~\ref{fig:coarse} show that the proposed approach can achieve almost identical performance with the baseline with various architectures and datasets. 

\subsection{Evaluation with Computation Time}
Furthermore, we evaluate the efficiency of the proposed approach from the perspective of computation time. Specifically, we compare the computation time of the proposed method and the Taylor expansion method in a single iteration of ranking and a complete run of ranking, pruning and fine-tuning, respectively. Since the time used in each run decreases as the pruning procedure goes on, comparing standard deviation is trivial. Therefore, we only report the average computation time in this set of experiments.

The experimental results are shown in Table \ref{tab:time}. In all scenarios, the ranking time of our proposed approach is negligibly short, i.e., less than $0.01$ second. Specifically, with AlexNet on CIFAR-10, our method only needs $0.002$ second for one run of ranking while the baseline needs $5.27$ seconds. Moreover, our ranking time increases slightly even though the model grows much deeper (from AlexNet to VGG-16) and the dataset becomes much larger (from CIFAR-10 to ImageNet). However, the corresponding computation time of the baseline method increases significantly. 
\begin{table}[htb]
\centering
\begin{tabular}{|c|c|c||c|c|}
\hline
Model & RT (ours) & RT (TE) & TT (ours) & TT (TE)  \\
\hline
A-C & 0.002 & 5.27 & 171 & 692\\
\hline
V-B & 0.005 & 40.18  & 3098 & 8027 \\
\hline
V-F & 0.007 & 18.89 & 1894 & 5023\\
\hline
V-I & 0.006 & 112.39 & 11458 & 24903\\
\hline
\end{tabular}
\vspace{-0.2cm}
\caption{Comparison of computation time (in seconds). RT: ranking time, TT: total pruning time. A-C: AlexNet on Cifar-10, V-B: VGG-16 on Birds-200, V-F: VGG-16 on Flowers-102, V-I: VGG-16 on ImageNet.}
\label{tab:time}
\vspace{-0.2cm}
\end{table}

For a complete run of ranking, pruning and fine-tuning, the total computation time of our approach is $171$ seconds while the baseline needs $692$ seconds with AlexNet on CIFAR-10. Our method reduces $75\%$ of the computation time, compared with the baseline. For VGG-16, our approach can reduce around $60\%$ of the overall computation time. The percentages of reduced time are different because in the fine-tuning phase, the number of updates varies according to the specific task. For VGG-16, more updates are required to achieve competitive performance.


\subsection{Correlation Analysis of Coarse and Precise Rankings}
\label{corr}
The goal of this section is to study the correlation between the coarse ranking results calculated with our approach and the precise ranking results with Taylor expansion. The Spearman's rank correlation coefficient described in Section \ref{existingchannelpruning} is utilized as the measurement. 

Firstly, we study the variation of two precise rankings with the same set of hyperparameters, and the results are shown in Table \ref{tab:correlation} (rightmost column). It is observed that the difference of two precise ranking is noticeable, especially with VGG-16 on ImageNet. This is because training samples are shuffled and are fed into the network with mini-batches.
This variation can be considered as a correlation baseline when we evaluate the correlation between precise ranking and coarse ranking in the following paragraph.

We report the correlation between the proposed coarse ranking and the precise ranking in different scenarios, fine-tuning with learning rate $10^{-5}$ and $10^{-4}$, respectively, shown in Table \ref{tab:correlation} (2nd and 3rd column). We can observe that the rank correlation between the proposed coarse ranking and precise ranking is close to the correlation baseline (two precise rankings). In specific, the correlation with smaller learning rate is bigger since smaller learning rate leads to smaller change of the network. Besides, the correlation on ImageNet is relatively smaller compared with other datasets, and the probable reason is that ImageNet is a huge dataset containing millions of samples, so it is challenging to achieve an appropriate ranking even with the precise ranking \cite{molchanov2016pruning}. 
\vskip -0.1in
\begin{table}[htb]
\centering
\begin{tabular}{|c|c|c|c|}
\hline
Model & Corr (1e-5) & Corr (1e-4) & Variation \\
\hline
A-C & 0.81 $\pm$ 0.17 & 0.30 $\pm$ 0.23 & 0.96 $\pm$ 0.06 \\
\hline
V-B & 0.42 $\pm$ 0.05 & 0.37 $\pm$ 0.04 & 0.78 $\pm$ 0.17 \\
\hline
V-F & 0.73 $\pm$ 0.19 & 0.46 $\pm$ 0.16 & 0.88 $\pm$ 0.17 \\
\hline
V-I & 0.32 $\pm$ 0.14 & 0.17 $\pm$ 0.07 & 0.69 $\pm$ 0.23 \\
\hline
\end{tabular}
\vspace{-0.2cm}
\caption{Correlation comparison.}
\label{tab:correlation}
\vspace{-0.2cm}
\end{table}
\subsection{Supplementary Verification with Mean Activation}
\begin{figure}[htb]
\centering
\vskip -0.15in
\begin{minipage}{\columnwidth}
\centering
\subfigure[AlexNet,CIFAR-10]{
\includegraphics[width=0.45\textwidth]{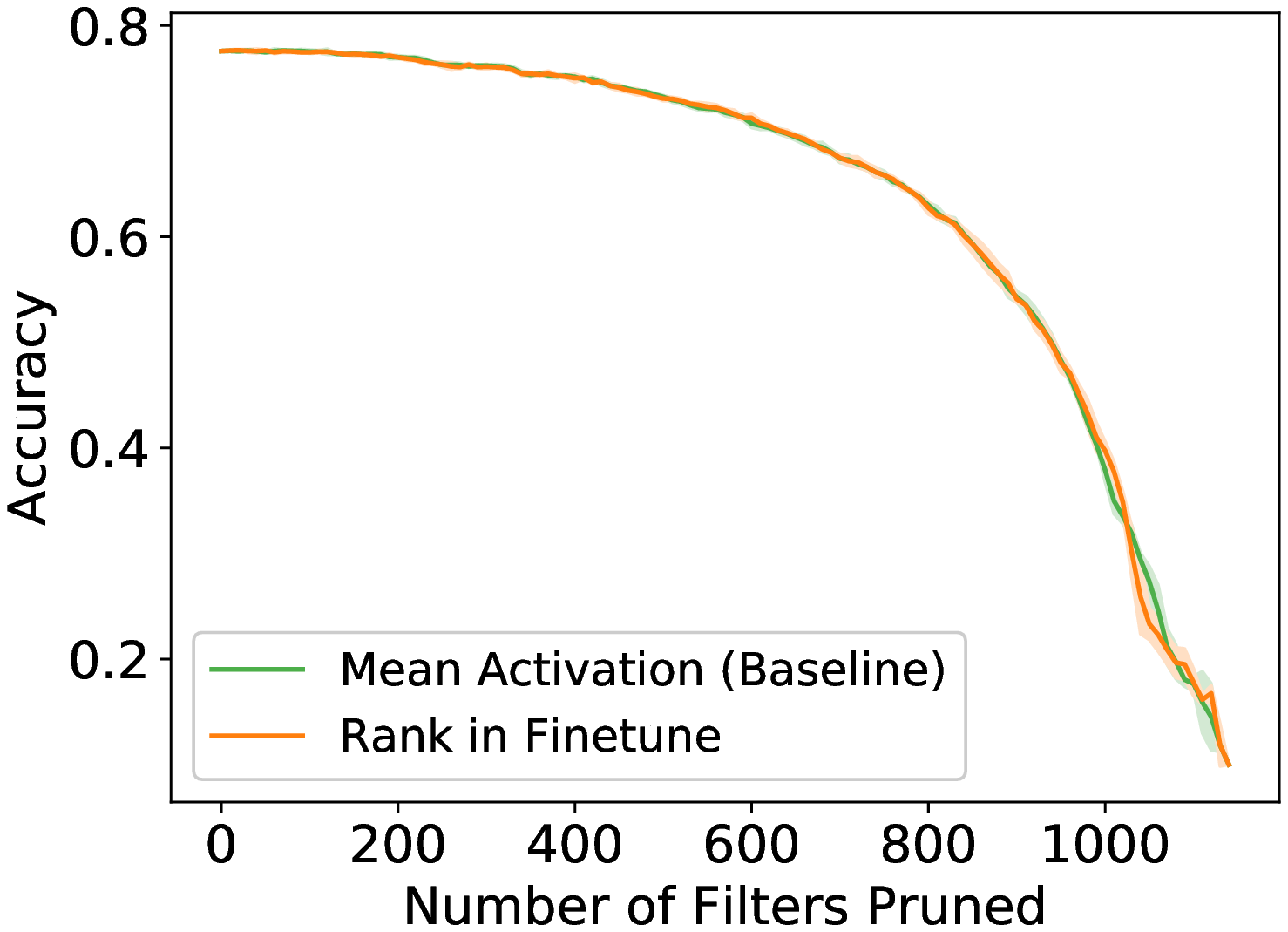}}
\subfigure[VGG-16,Flowers]{
\includegraphics[width=0.45\textwidth]{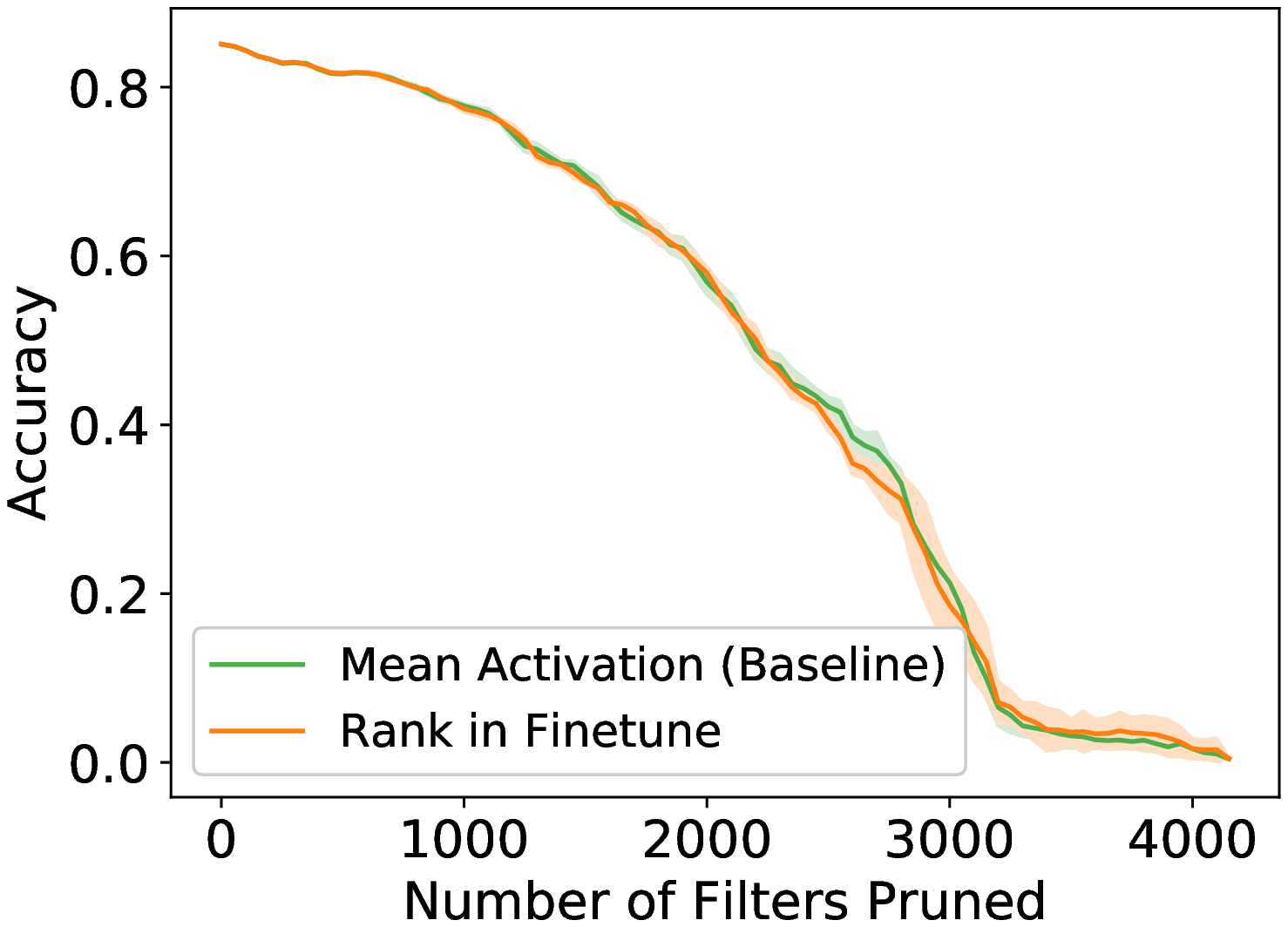}}
\end{minipage}
\vskip -0.15in
\caption{Performance comparison of the proposed approach (Rank in Finetune) and its corresponding baseline (mean activation).}
\label{fig:mean_act}
\vskip -0.28in
\end{figure}

To verify the generality of our proposed methodology, we evaluate the proposed approach with another widely-used channel pruning approach, i.e., the mean activation approach. 
Experimental results on classification accuracy are shown in Fig.~\ref{fig:mean_act}. Once again, it is observed that our approach  achieves almost identical performance with the baseline while our computation time for ranking is negligibly shot since we do not need to redo feed-forward process in a separate phrase.
\section{Conclusion}
\label{sec:con}
In this paper, we proposed a novel channel pruning framework that integrates the ranking phase and the fine-tuning phase by sharing intermediate computation results. Extensive experiments showed that our approach can significantly reduce the ranking time while achieving almost identical classification accuracy with the state-of-the-art channel pruning methods with data-driven filter ranking criteria. The proposed approach would significantly facilitate the pruning practice, especially on resource-constrained platforms.

\bibliographystyle{IEEEbib}
\bibliography{refs}

\begin{thebibliography}{10}

\bibitem{iandola2014densenet}
Forrest Iandola, Matt Moskewicz, Sergey Karayev, Ross Girshick, Trevor Darrell,
  and Kurt Keutzer,
\newblock ``Densenet: Implementing efficient convnet descriptor pyramids,''
\newblock {\em arXiv preprint arXiv:1404.1869}, 2014.

\bibitem{li2018fast}
Chengcheng Li, Zi~Wang, and Hairong Qi,
\newblock ``Fast-converging conditional generative adversarial networks for
  image synthesis,''
\newblock {\em arXiv preprint arXiv:1805.01972}, 2018.

\bibitem{christiansen2018silico}
Eric~M Christiansen, Samuel~J Yang, D~Michael Ando, Ashkan Javaherian, Gaia
  Skibinski, Scott Lipnick, Elliot Mount, Alison O’Neil, Kevan Shah, Alicia~K
  Lee, et~al.,
\newblock ``In silico labeling: Predicting fluorescent labels in unlabeled
  images,''
\newblock {\em Cell}, vol. 173, no. 3, pp. 792--803, 2018.

\bibitem{Yuanchaodaen}
Yuanchao Su, Jun Li, Antonio Plaza, Andrea Marinoni, Paolo Gamba, and Somdatta
  Chakravortty,
\newblock ``Daen: Deep autoencoder networks for hyperspectral unmixing,''
\newblock {\em IEEE Transactions on Geoscience and Remote Sensing}, 2019.

\bibitem{wang2018deep}
Zi~Wang, Dali Wang, Chengcheng Li, Yichi Xu, Husheng Li, and Zhirong Bao,
\newblock ``Deep reinforcement learning of cell movement in the early stage of
  c. elegans embryogenesis,''
\newblock {\em arXiv preprint arXiv:1801.04600}, 2018.

\bibitem{wang2019cellular}
Dali Wang, Zheng Lu, Yichi Xu, Zi~Wang, Chengcheng Li, Anthony Santella, and
  Zhirong Bao,
\newblock ``Cellular structure image classification with small targeted
  training samples,''
\newblock {\em bioRxiv}, p. 544130, 2019.

\bibitem{han2015learning}
Song Han, Jeff Pool, John Tran, and William Dally,
\newblock ``Learning both weights and connections for efficient neural
  network,''
\newblock in {\em Advances in neural information processing systems}, 2015, pp.
  1135--1143.

\bibitem{li2016pruning}
Hao Li, Asim Kadav, Igor Durdanovic, Hanan Samet, and Hans~Peter Graf,
\newblock ``Pruning filters for efficient convnets,''
\newblock {\em arXiv preprint arXiv:1608.08710}, 2016.

\bibitem{molchanov2016pruning}
Pavlo Molchanov, Stephen Tyree, Tero Karras, Timo Aila, and Jan Kautz,
\newblock ``Pruning convolutional neural networks for resource efficient
  inference,''
\newblock {\em arXiv preprint arXiv:1611.06440}, 2016.

\bibitem{liu2017learning}
Zhuang Liu, Jianguo Li, Zhiqiang Shen, Gao Huang, Shoumeng Yan, and Changshui
  Zhang,
\newblock ``Learning efficient convolutional networks through network
  slimming,''
\newblock in {\em Computer Vision (ICCV), 2017 IEEE International Conference
  on}. IEEE, 2017, pp. 2755--2763.

\bibitem{he2017channel}
Yihui He, Xiangyu Zhang, and Jian Sun,
\newblock ``Channel pruning for accelerating very deep neural networks,''
\newblock in {\em International Conference on Computer Vision (ICCV)}, 2017,
  vol.~2.

\bibitem{polyak2015channel}
Adam Polyak and Lior Wolf,
\newblock ``Channel-level acceleration of deep face representations,''
\newblock {\em IEEE Access}, vol. 3, pp. 2163--2175, 2015.

\bibitem{han2016eie}
Song Han, Xingyu Liu, Huizi Mao, Jing Pu, Ardavan Pedram, Mark~A Horowitz, and
  William~J Dally,
\newblock ``Eie: efficient inference engine on compressed deep neural
  network,''
\newblock in {\em Computer Architecture (ISCA), 2016 ACM/IEEE 43rd Annual
  International Symposium on}. IEEE, 2016, pp. 243--254.

\bibitem{hu2016network}
Hengyuan Hu, Rui Peng, Yu-Wing Tai, and Chi-Keung Tang,
\newblock ``Network trimming: A data-driven neuron pruning approach towards
  efficient deep architectures,''
\newblock {\em arXiv preprint arXiv:1607.03250}, 2016.

\bibitem{zar1972significance}
Jerrold~H Zar,
\newblock ``Significance testing of the spearman rank correlation
  coefficient,''
\newblock {\em Journal of the American Statistical Association}, vol. 67, no.
  339, pp. 578--580, 1972.

\bibitem{krizhevsky2012imagenet}
Alex Krizhevsky, Ilya Sutskever, and Geoffrey~E Hinton,
\newblock ``Imagenet classification with deep convolutional neural networks,''
\newblock in {\em Advances in neural information processing systems}, 2012, pp.
  1097--1105.

\bibitem{krizhevsky2009learning}
Alex Krizhevsky and Geoffrey Hinton,
\newblock ``Learning multiple layers of features from tiny images,''
\newblock Tech. {R}ep., Citeseer, 2009.

\bibitem{simonyan2014very}
Karen Simonyan and Andrew Zisserman,
\newblock ``Very deep convolutional networks for large-scale image
  recognition,''
\newblock {\em arXiv preprint arXiv:1409.1556}, 2014.

\bibitem{WelinderEtal2010}
P.~Welinder, S.~Branson, T.~Mita, C.~Wah, F.~Schroff, S.~Belongie, and
  P.~Perona,
\newblock ``{Caltech-UCSD Birds 200},''
\newblock Tech. {R}ep. CNS-TR-2010-001, California Institute of Technology,
  2010.

\bibitem{nilsback2008automated}
Maria-Elena Nilsback and Andrew Zisserman,
\newblock ``Automated flower classification over a large number of classes,''
\newblock in {\em Computer Vision, Graphics \& Image Processing, 2008.
  ICVGIP'08. Sixth Indian Conference on}. IEEE, 2008, pp. 722--729.

\bibitem{deng2009imagenet}
Jia Deng, Wei Dong, Richard Socher, Li-Jia Li, Kai Li, and Li~Fei-Fei,
\newblock ``Imagenet: A large-scale hierarchical image database,''
\newblock in {\em Computer Vision and Pattern Recognition, 2009. CVPR 2009.
  IEEE Conference on}. Ieee, 2009, pp. 248--255.

\bibitem{paszke2017automatic}
Adam Paszke, Sam Gross, Soumith Chintala, Gregory Chanan, Edward Yang, Zachary
  DeVito, Zeming Lin, Alban Desmaison, Luca Antiga, and Adam Lerer,
\newblock ``Automatic differentiation in pytorch,''
\newblock 2017.

\end{thebibliography}
\end{document}